\documentclass[journal,transmag]{IEEEtran}
\ifCLASSINFOpdf
\else
\fi

\usepackage{times}  
\usepackage{helvet} 
\usepackage{courier}  
\usepackage[hyphens]{url}  
\usepackage{graphicx} 

\usepackage{latexsym}

\usepackage[utf8]{inputenc}
\usepackage[small]{caption}
\usepackage{amsmath}
\usepackage{booktabs}
\usepackage{algorithm}
\usepackage{algorithmic}
\usepackage{url,ifthen}
\usepackage{amssymb}

\begin{document}
%
\title{Mobile Networks for Computer Go}


\author{\IEEEauthorblockN{Tristan Cazenave}
\IEEEauthorblockA{LAMSADE, Université Paris-Dauphine, PSL, PRAIRIE, CNRS, Paris, France}
\thanks{Corresponding author: T. Cazenave (email: Tristan.Cazenave@dauphine.psl.eu)}}

%



\IEEEtitleabstractindextext{%
\begin{abstract}
The architecture of the neural networks used in Deep Reinforcement Learning programs such as Alpha Zero or Polygames has been shown to have a great impact on the performances of the resulting playing engines. For example the use of residual networks gave a 600 ELO increase in the strength of Alpha Go. This paper proposes to evaluate the interest of Mobile Network for the game of Go using supervised learning as well as the use of a policy head and a value head different from the Alpha Zero heads. The accuracy of the policy, the mean squared error of the value, the efficiency of the networks with the number of parameters, the playing speed and strength of the trained networks are evaluated.
\end{abstract}

}

\maketitle

\IEEEdisplaynontitleabstractindextext

%
\IEEEpeerreviewmaketitle

\section{Introduction}

This paper is about the efficiency of neural networks trained to play the game of Go. Mobile Networks \cite{howard2017mobilenets,sandler2018mobilenetv2} are commonly used in computer vision to classify images. They obtain high accuracy for standard computer vision datasets while keeping the number of parameters lower than other neural networks architectures.

In computer Go and more generally in board games the neural networks usually have more than one head. They have at least a policy head and a value head. The policy head is evaluated with the accuracy of predicting the moves of the games and the value head is evaluated with the Mean Squared Error (MSE) on the predictions of the outcomes of the games. The current state of the art for such networks is to use residual networks \cite{Cazenave2018residual,silver2017mastering,silver2018general}.

The architectures used for neural networks in supervised learning and Deep Reinforcement Learning in games can greatly change the performances of the associated game playing programs. For example residual networks gave AlphaGo Zero a 600 ELO gain in playing strength compared to standard convolutional neural networks.

Residual networks will be compared to Mobile Networks for computer Go. Different options for the policy head and the value head will also be compared. The basic residual networks used for comparison are networks following exactly the AlphaGo Zero and Alpha Zero architectures. The improvements due to Mobile Networks and changes in the policy head and the value head are not specific to computer Go and can be used without modifications for other games.

This research originated from a computer Go tournament I organized for my master students. In order for the students to validate my Deep Learning course for the IASD master at University Paris-Dauphine, PSL, I made them train computer Go neural networks they could submit to tournaments played on a weekly basis \cite{projetdeep}. In order to be fair about training resources the number of parameters for the networks had to be lower than 1 000 000. The goal was to train a neural network using supervised learning on a dataset of 500 000 games played by ELF/OpenGo \cite{tian2019elfopengo} at a superhuman level. There were more than 109 000 000 different states in this dataset. I gave the students a Python library I programmed in C++ so as to randomly build batches of tensors representing states that could be used to give inputs and outputs to the networks. I also programmed a Monte Carlo Tree Search (MCTS) algorithm close to PUCT so as to make the students neural networks play against each other in a round robin tournament.

The remainder of the paper is organized as follows. The second section presents related works in Deep Reinforcement Learning for games. The third section describes the training and the test sets. The fourth section details the neural networks that are tested for the game of Go. The fifth section gives experimental results.

\section{Zero Learning}

Monte Carlo Tree Search (MCTS) \cite{Coulom2006,Kocsis2006} made a revolution in board games Artificial Intelligence. A second revolution occurred when it was combined with Deep Reinforcement Learning which led to superhuman level of play in the game of Go with the AlphaGo program \cite{silver2016mastering}.

Residual networks \cite{Cazenave2018residual}, combined with policy and value sharing the same network and Expert Iteration \cite{anthony2017thinking} did improve much on AlphaGo leading to AlphaGo Zero \cite{silver2017mastering} and zero learning. With these improvements AlphaGo Zero was able to learn the game of Go from scratch and surpassed AlphaGo.

Later Alpha Zero successfully applied the same principles to the games of Chess and Shogi \cite{silver2018general}.

Other researchers developed programs using zero learning to play various games.

ELF/OpenGo \cite{tian2019elfopengo} is an open-source implementation of AlphaGo Zero for the game of Go. After two weeks of training on 2 000 GPUs it reached superhuman level and beat professional Go players.

Leela Zero \cite{pascutto2017leela} is an open-source program that uses a community of contributors who donate GPU time to replicate the Alpha Zero approach. It has been applied with success to Go and Chess.

Crazy Zero by Rémi Coulom is a zero learning framework that has been applied to the game of Go as well as Chess, Shogi, Gomoku, Renju, Othello and Ataxx. With limited hardware it was able to reach superhuman level at Go using large batches in self-play and improvements of the targets to learn such as learning territory in Go. Learning territory in Go increases considerably the speed of learning.

KataGo \cite{greatfast} is an open-source implementation of AlphaGo Zero that improves learning in many ways. It converges to superhuman level much faster than alternative approaches such as Elf/OpenGo or Leela Zero. It makes use of different optimizations such as using a low number of playouts for most of the moves in a game so as to have more data about the value in a shorter time. It also uses additional training target so as to regularize the networks.

Galvanise Zero \cite{gzero} is an open-source program that is linked to General Game Playing (GGP) \cite{Pitrat68}. It uses rules of different games represented in the Game Description Language (GDL) \cite{love2008general}, which makes it a truly general zero learning program able to be applied as is to many different games. The current games supported by Galvanise Zero are Chess, Connect6, Hex11, Hex13, Hex19, Reversi8, Reversi10, Amazons, Breakthrough, International Draughts.

Polygames \cite{cazenave2020polygames} is a generic implementation of Alpha Zero that has been applied to many games, surpassing human players in difficult games such as Havannah and using architectural innovations such as a fully convolutional policy head.

\section{The Training and the Test Sets}

We use two datasets for training the networks.

The first dataset used for training comes from the Leela Zero Go program self played games. The selected games are the last 2 000 000 games of self play, starting at game number 19 000 000. The input data is composed of 21 19x19 planes (color to play, ladders, liberties, current state on two planes, four previous states on four planes). The output targets are the policy (a vector of size 361 with 1.0 for the move played, 0.0 for the other moves), the value (1.0 if White won, 0.0 if Black won).

The second dataset is the ELF dataset. It is built from the last 1 347 184 games played by ELF, it contains 301 813 318 states.


At the beginning of training and for each dataset 100 000 games are taken at random as a validation set and one state is selected for each game to be included in the validation set. The validation set for the Leela dataset only contains games from Leela and the validation set for the ELF dataset only contains games from ELF. The same set of states in the validation sets are used for all networks. These games and states are never used for training, none of the states present in the same game as a state in the test set are used for training. We define one epoch as 1 000 000 samples. For each sample in the training set a random symmetry among the eight possible symmetries is chosen.

Both datasets contain games played at superhuman level. The Leela games are played at a better level than the ELF games since the latest versions of Leela are stronger than ELF.


\section{Networks Architectures, Training and Use}

\subsection{Residual Networks}

Residual networks improve much on convolutional networks for the game of Go \cite{Cazenave2018residual,silver2017mastering}. In AlphaGo Zero they gave an increase of 600 ELO in the level of play. The principle of residual networks is to add the input of a residual block to its output. A residual block is composed of two convolutional layers with ReLU activations and batch normalization. For our experiments we use for Alpha Zero like networks the same block as in AlphaGo Zero.

Another architecture optimization used in AlphaGo Zero is to combine the policy and the value in a single network with two heads. It also enables an increase of 600 ELO in the level of play \cite{silver2017mastering}. All the networks we test have two heads, one for the policy and one for the value.

\subsection{Mobile Networks}

MobileNet \cite{howard2017mobilenets} followed by MobileNetV2 \cite{sandler2018mobilenetv2} provide a parameter efficient neural network architecture for computer vision. The principle of MobileNetV2 is to have blocks as in residual networks where the input of a block is added to its output. But instead of usual convolutional layers in the block they use depthwise convolutions. Moreover the number of channels at the input and the output of the blocks (in the trunk) is much smaller than the number of channels for the depthwise convolutions in the block. In order to efficiently pass from a small number of channels in the trunk to a greater number in the block, usual convolutions with cheap 1x1 filters are used at the entry of the block and at its output.

The Keras \cite{chollet2015keras,chollet2017deep} source code we used for the Mobile models is given in the appendix.

\subsection{Optimizing the Heads}

The AlphaGo Zero policy head uses 1x1 convolutions to project the 256 channels to two channels and then it flattens the channels and uses a dense layer with 362 outputs for all possible legal moves in Go. The AlphaGo Zero value head uses 1x1 convolutions to project the 256 channels to one channel and then it flattens the channel, connects it to a dense layer with 256 outputs and then connects these outputs to a single output for the value \cite{silver2017mastering}.

We experimented with different policy and value heads. For the policy head we tried a fully convolutional policy head. It does not use a dense layer. Instead it uses 1x1 convolutions to project the channels to a single channel, then it simply flattens the channel directly giving 361 outputs, one for each possible move except the pass move. The fully convolutional head has already been used in Polygames \cite{cazenave2020polygames}.

For the value head we experimented with average pooling. The use of Spatial Average Pooling in the value head has already been shown to be an improvement for Golois \cite{Cazenave18sap}. It was also used in Katago \cite{greatfast} and in Polygames \cite{cazenave2020polygames}. In this paper we experiment with Global Average Pooling for the value head. Each channel is averaged among its whole 19x19 plane leading to a vector of size equal to the number of channels. It is then connected to a dense layer with 50 outputs. The last layer is a dense layer with one output for the value.

\subsection{Training}

Training of the networks uses the Keras/Tensorflow framework. We define an epoch as 1 000 000 states. The evaluation on the test set is computed every epoch. The loss used for the value in the Alpha Zero papers is the mean squared error (MSE). We keep this loss for the validation and the tests of the networks in order to compare them on an equal basis. In some of the network we train the value with the binary cross entropy loss which seems more adapted to the learning of the value (i.e. we want to know if the game is won or lost). We also experiment with a weight on the value loss. The binary cross entropy loss is usually greater than the mean squared error loss, but we can make it even greater by multiplying the loss with a constant.

The batch size is fixed to 32. The annealing schedule is to train with a learning rate of 0.005 for the first 100 epochs. Then to train with a learning rate of 0.0005 from 100 to 150 epochs. Then to train with a learning rate of 0.00005 from 150 to 200 epochs. It enables to fine tune the networks when the learning stalls. This is similar to the Alpha Zero annealing schedule which also divides the learning rate by ten every 200 epochs in the beginning and every 100 epochs in the end. Using this schedule the training of a large mobile network approximately takes 12 days with a V100 card.

For all networks we use a L2 regularization during training with a weight of 0.0001. We found that the validation loss and the level of the trained network is much better when using regularization.


\subsection{Inputs and Outputs}

The inputs of the networks use the colors of the stones, the liberties, the ladder status of the stone, the ladder status of adjacent strings (i.e. if an adjacent string is in ladder), the last 5 states and a plane for the color to play. The total number of planes used to encode a state is 21.

The outputs are a 0 or a 1 for the value head. A 0 means Black has won the game and a 1 means it is White. For the policy head there is a 1 for the move played in the state and 
0 for all other moves. The output for the policy head is different from the output used in Alpha Zero since AlphaGo Zero and Alpha Zero use Expert Iteration \cite{anthony2017thinking} which gives as output the number of time the moves has been tried in the PUCT search divided by the total number of evaluations in the PUCT search.

\subsection{Self Play Speed}

A program that plays games against itself so as to generate more training data can be strongly parallelized. Parallelizing the different games being played can greatly speedup the overall reinforcement learning process. Both the forward pass of the network and the building of the batches can be parallelized. Parallelizing the forward passes is effectively done by building large batches of states with one state per self played game. The GPU is good at effectively parallelizing the forward pass on large batches. The building of the inputs of the large batches can also be strongly parallelized using threads.

Smaller networks are faster and enable larger batches for self play. This is why most programs start training with small networks and make them grow during learning.

\section{Experimental Results}

For all the experiments the training uses batch of 256 samples. The learning rate starts at 0.005 for the whole batch as it is a stable learning rate that decreases the loss as fast as possible. The learning rate is then divided by 10 at 100 and 150 epochs in order to fine tune the networks. The regularizer is the L2 loss with a weight of 0.0001.

We had problems with the Alpha Zero value head: it often did not learn even after many epochs so we replace it with another value head using average pooling. The use of average pooling layers for the value has been described previously in Golois \cite{Cazenave18sap}, KataGo \cite{greatfast} and Polygames \cite{cazenave2020polygames}. The value head we used has a global average pooling layer followed by a dense layer of 50 neurons and another dense layer with one output. We used the same value head for all our networks since it gave better results than the Alpha Zero value head. Even with this value head it was necessary to launch multiple times the training of the large Alpha Zero like networks in order to start the convergence of the value.

\subsection{Networks with less than one million parameters}

The Alpha Zero like network has 10 residual blocks of 63 filters and the Alpha Zero policy head. It has 986 748 parameters. During training it uses the MSE loss for the value and the Categorical Crossentropy loss for the policy. The network is called a0.small.

The Alpha Zero like fully convolutional network has 13 residual blocks of 64 filters. For the policy head it does not use a dense layer, just a 1x1 convolution to a single plane and a flatten. The usual residual blocks used by Alpha Zero can have problems with this policy head (the policy loss initially stays close to zero). It is better to use the Golois residual blocks \cite{cazenave2017improved}: the rectifier is after the convolution, the batch normalization is after the addition. It has 968 485 parameters. During training it uses the Binary Crossentropy loss for the value and the Categorical Crossentropy loss for the policy. The network is called a0.small.conv.bin.

The third network is the same as the Alpha Zero like fully convolutional network except that it uses a weight of 4 for the value loss. It is called a0.small.conv.bin.val4.

The fourth network uses 25 MobileNet blocks with a trunk of 64 and 200 filters inside the blocks. It uses the Alpha Zero policy head. It has 997 506 parameters. During training it uses the MSE loss for the value and the Categorical Crossentropy loss for the policy. The network is called mobile.small.

The fifth network uses 33 MobileNet blocks with a trunk of 64 and 200 filters inside the blocks. It uses the fully convolutional policy head. It has 970 477 parameters. During training it uses the Binary Crossentropy loss for the value and the Categorical Crossentropy loss for the policy. The network is called mobile.small.conv.bin.

The sixth network is the same as the fifth network except that it has a weight of 4 on the value loss. The network is called mobile.small.conv.bin.val4.

\begin{figure}
\centering
\includegraphics[scale=0.5]{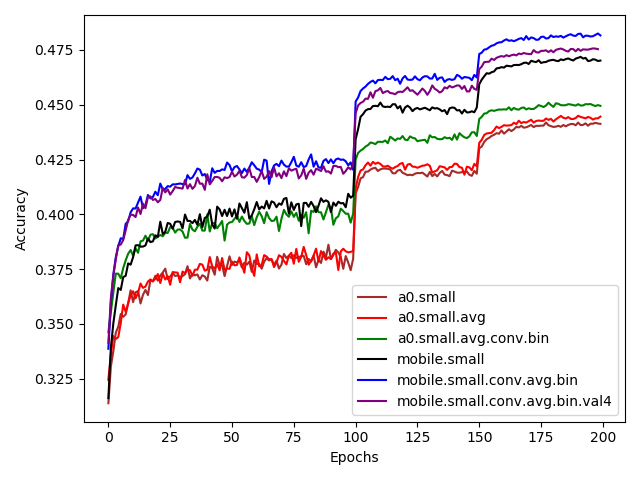}
\caption{The evolution of the policy validation accuracy for the different networks with less than one million parameters.}
\label{smallPolicy}
\end{figure}

Figure \ref{smallPolicy} gives the evolution of the accuracy for all small networks. The Alpha Zero like networks have a lower accuracy than the Mobile networks. The best network use MobileNet blocks together with a fully convolutional policy head and global average pooling for the value head. The Alpha Zero like network has the worst results. When removing the policy head and keeping only a 1x1 convolution the results get better. Using MobileNets with the Alpha Zero policy head is close to the fully convolutional Alpha Zero network. Training a fully convolutional MobileNet improves much the results. Finally putting a weight of four on the value loss of the fully convolutional MobileNet does not hurt much the training of the policy.

\begin{figure}
\centering
\includegraphics[scale=0.5]{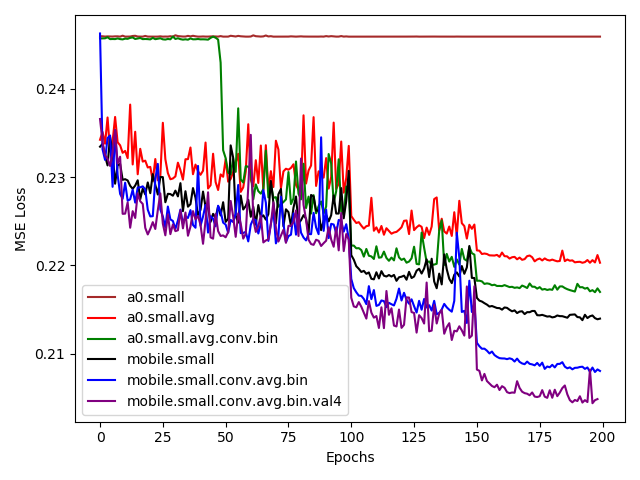}
\caption{The evolution of the value validation MSE loss for the different networks with less than one million parameters.}
\label{smallValue}
\end{figure}

We can see in figure \ref{smallValue} that the small Alpha Zero like network does not learn the value within 200 epochs. We tried to launch the a0 training multiple times but did not succeed in learning both the policy and the value with a small network on the Leela dataset. The best value is obtained with a MobileNet with a weight of 4 on the value loss. With a weight of 1 the MobileNet is still the second best for the value, better than the Alpha Zero like networks.


\subsection{Unbounded Networks}

We now experiments with large networks of sizes similar to the sizes of the Alpha Zero networks. 

The Alpha Zero like networks have $n$ residual blocks of 256 filters and the Alpha Zero policy head. During training they uses the MSE loss for the value and the Categorical Crossentropy loss for the policy. The networks are called a0.n. We test the networks with 10, 20 and 40 residual blocks.

The MobileNets networks use $n$ MobileNet blocks with a trunk of 128 and 512 filters inside the blocks. They use the Alpha Zero policy head. During training they use the MSE loss for the value and the Categorical Crossentropy loss for the policy. The network are called mobile.n. We test the networks with 10, 20 and 40 MobileNet blocks.

The MobileNets fully convolutional networks use $n$ MobileNet blocks with a trunk of 128 and 512 filters inside the blocks. They use the fully convolutional policy head. During training they use the Binary Crossentropy loss for the value and the Categorical Crossentropy loss for the policy. The networks are called mobile.conv.bin.n. We test the networks with 10, 20 and 40 MobileNet blocks.

The MobileNets fully convolutional networks with a weight of 4 on the value loss are called mobile.conv.bin.val4.n. We test the networks with 10, 20 and 40 MobileNet blocks.

\begin{figure}
\centering
\includegraphics[scale=0.5]{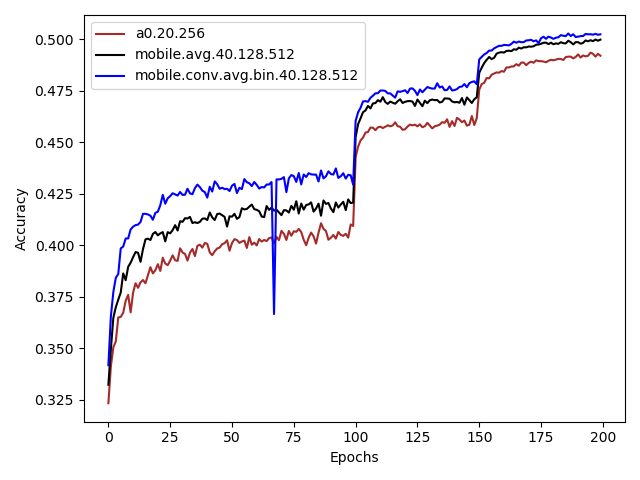}
\caption{The evolution of the validation policy accuracy for the different unbounded networks.}
\label{policy}
\end{figure}

\begin{figure}
\centering
\includegraphics[scale=0.5]{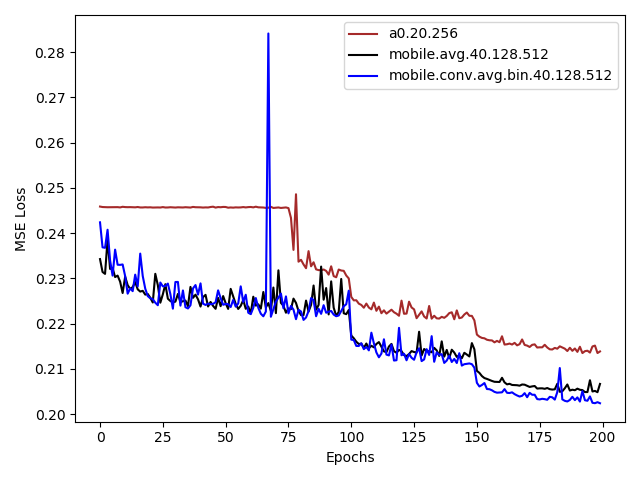}
\caption{The evolution of the validation value MSE loss for the different unbounded networks.}
\label{value}
\end{figure}

Figure \ref{policy} gives the validation policy accuracy for the Alpha Zero like network and two Mobile networks. The Mobile networks have better accuracy and the fully convolutional policy head is slightly better.

Figure \ref{value} show that the validation MSE loss of the value is also better for Mobile networks than for Alpha Zero like networks. 


\subsection{Parameter Efficiency}

\begin{figure}
\centering
\includegraphics[scale=0.5]{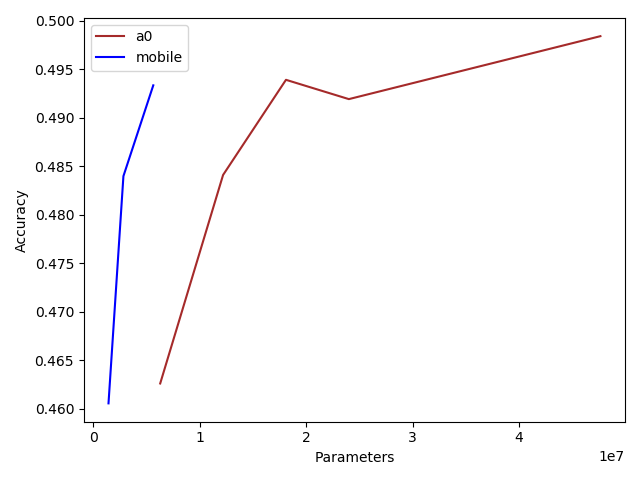}
\caption{The evolution of the policy validation accuracy with the number of parameters.}
\label{policyPareto}
\end{figure}

\begin{figure}
\centering
\includegraphics[scale=0.5]{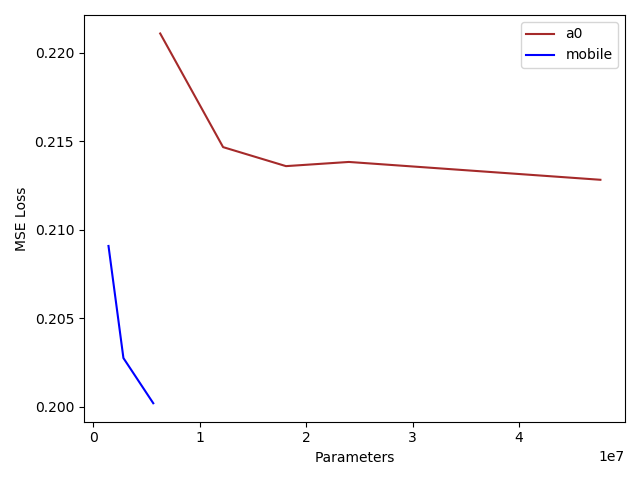}
\caption{The evolution of the value validation MSE Loss with the number of parameters.}
\label{valuePareto}
\end{figure}

We now give results for the validation accuracy and the validation MSE loss according to the number of parameters of the networks. We compare Mobile networks with fully convolutional policy head and global average pooling value head to Alpha Zero residual networks.

Figure \ref{policyPareto} gives the accuracy of the different networks according to the number of parameters. The Mobile networks that are trained have 10, 20 and 40 Mobile blocks, a trunk of 128 and 512 filters inside the blocks. The Alpha Zero networks have 5, 10, 15, 20, 30 and 40 residual blocks of 256 filters. Mobile networks have a comparable accuracy with much less parameters


Figure \ref{valuePareto} gives the MSE loss of Mobile networks and residual network according to the number of parameters. Mobile networks have a much better evaluation than residual networks with much fewer parameters.

\subsection{Training on ELF self-played games}

Learning the value is difficult for Alpha Zero like networks on the Leela games. This may be due to Leela Zero resigning long before the endgame in states difficult to evaluate. The ELF self-played games are from a weaker engine and contains states easier to evaluate. The same networks as in the previous section are tested on the ELF dataset.

\begin{figure}
\centering
\includegraphics[scale=0.5]{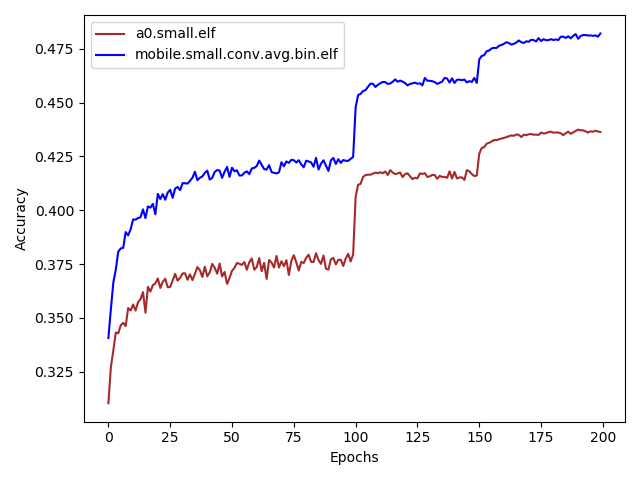}
\caption{The evolution of the policy validation accuracy for the different networks with less than one million parameters on the ELF dataset.}
\label{smallPolicyElf}
\end{figure}

We can see in figure \ref{smallPolicyElf} that the Alpha Zero like network is worse than a fully convolutional MobileNet on the ELF dataset with a network of less than 1 000 000 parameters.

\begin{figure}
\centering
\includegraphics[scale=0.5]{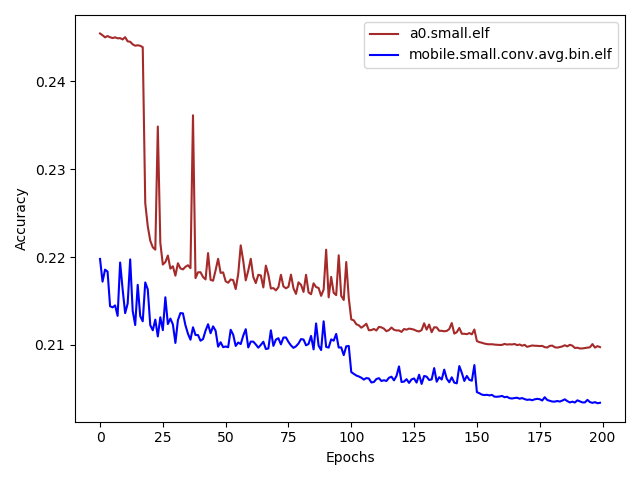}
\caption{The evolution of the value validation MSE loss for the different networks with less than one million parameters on the ELF dataset.}
\label{smallValueElf}
\end{figure}

Figure \ref{smallValueElf} shows that small Alpha Zero like networks can learn the value of the ELF dataset when they could not on the Leela dataset. Nevertheless, the small Mobile networks still better learn the value than the Alpha Zero like networks.

\begin{figure}
\centering
\includegraphics[scale=0.5]{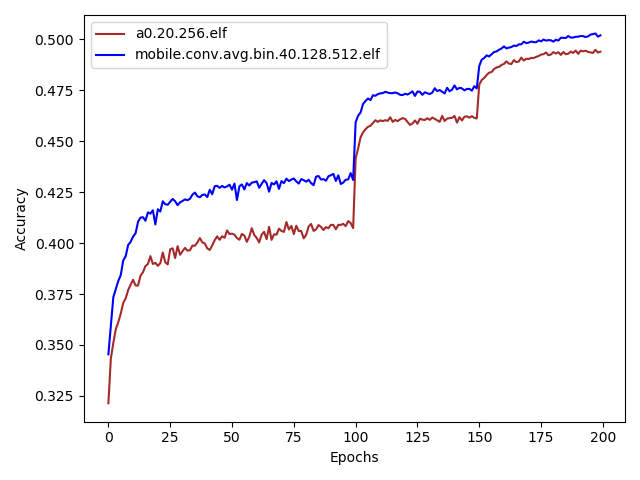}
\caption{The evolution of the validation policy accuracy for the different unbounded networks on the ELF dataset.}
\label{policyElf}
\end{figure}

We can see in figure \ref{policyElf} that large Mobile networks have a better policy accuracy that large Alpha Zero like networks even if the Mobile network tested has much less parameters than the Alpha Zero network.

\begin{figure}
\centering
\includegraphics[scale=0.5]{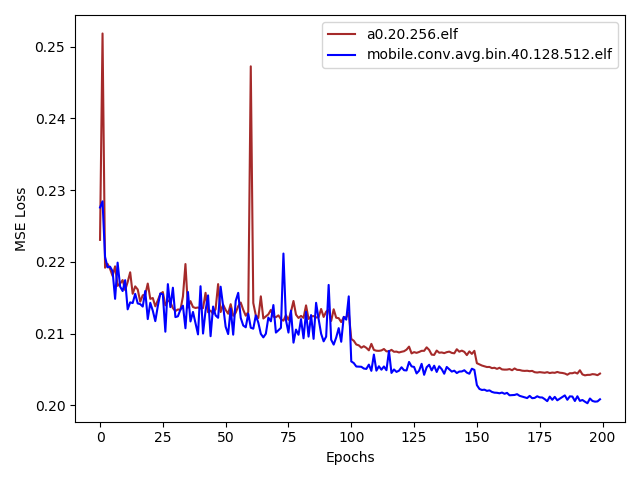}
\caption{The evolution of the validation value MSE loss for the different unbounded networks on the ELF dataset.}
\label{valueElf}
\end{figure}

Figure \ref{valueElf} show that the Mobile network we tested is slightly better for learning the value than the 20 blocks residual networks.

\begin{figure}
\centering
\includegraphics[scale=0.5]{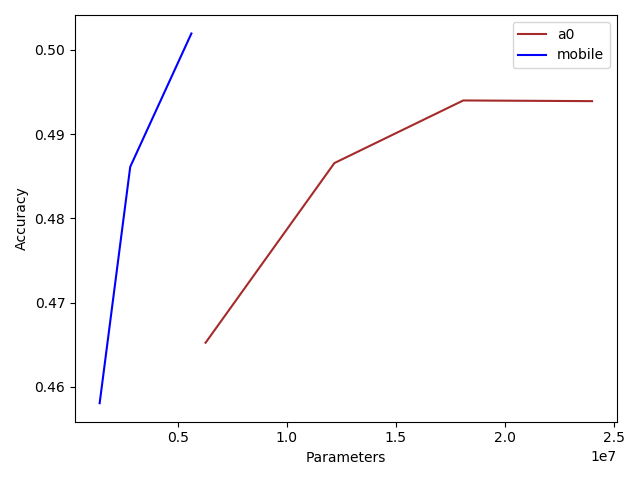}
\caption{The evolution of the policy validation accuracy with the number of parameters on the ELF dataset.}
\label{policyParetoElf}
\end{figure}

\begin{figure}
\centering
\includegraphics[scale=0.5]{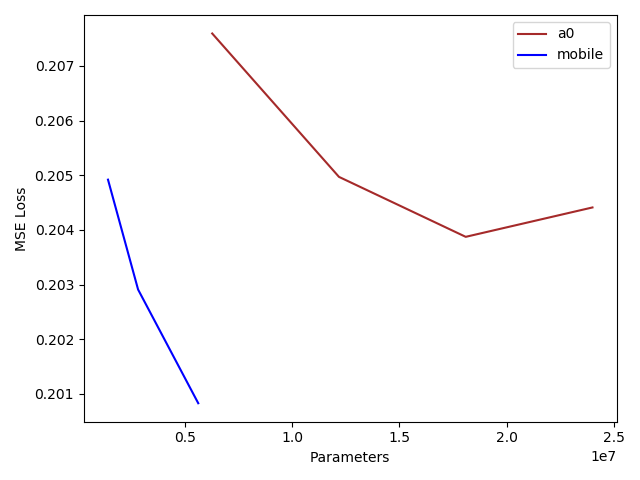}
\caption{The evolution of the value validation MSE Loss with the number of parameters on the ELF dataset.}
\label{valueParetoElf}
\end{figure}

Figure \ref{policyParetoElf} and \ref{valueParetoElf} show the parameter efficiency of Mobile and residual networks for the policy and the value on the ELF dataset. The policy accuracy and the value MSE loss are better for Mobile networks than for residual networks while using much less parameters. The networks used for this experiment are the 10, 20 and 40 blocks Mobile networks and the 5, 10, 15 and 20 residual blocks networks.

\subsection{Self Play Speed}

We can see in table \ref{tableSpeed} the number of states evaluated per second according to the size of the batch in input of the networks. For small batches residual networks are more than twice as fast as Mobile networks. For large batches residual networks are still faster but close to the speed of Mobile networks. The GPU used for the experiments is a RTX 2080 Ti and the CPU is a 24 cores computer.

\begin{table}
  \centering
  \caption{Comparison of the number of states per second.}
  \label{tableSpeed}
  \begin{tabular}{lrrrrrrrrrr}
  
  Network                        & Batch Size & Hardware & Speed  \\
                                 &            &                    \\
  a0.20.256                      &      4 096 &      GPU & 1 053.04 \\
  a0.20.256                      &      8 192 &      GPU & 1 598.43 \\
  a0.20.256                      &     16 384 &      GPU & 2 114.54 \\
  a0.20.256                      &     32 768 &      GPU & 2 533.68 \\
  a0.20.256                      &     65 536 &      GPU & 2 536.46 \\
  mobile.conv.avg.bin.40.128.512 &      4 096 &      GPU &   462.54 \\
  mobile.conv.avg.bin.40.128.512 &      8 192 &      GPU &   846.67 \\
  mobile.conv.avg.bin.40.128.512 &     16 384 &      GPU & 1 327.23 \\
  mobile.conv.avg.bin.40.128.512 &     32 768 &      GPU & 1 899.46 \\
  mobile.conv.avg.bin.40.128.512 &     65 536 &      GPU & 2 061.72 \\
  a0.20.256                      &         16 &      CPU &    15.88 \\
  a0.20.256                      &         32 &      CPU &    32.91 \\
  a0.20.256                      &         64 &      CPU &    64.64 \\
  a0.20.256                      &        128 &      CPU &   127.63 \\
  a0.20.256                      &        256 &      CPU &   247.54 \\
  a0.20.256                      &        512 &      CPU &   481.34 \\
  a0.20.256                      &      1 024 &      CPU &   995.14 \\
  a0.20.256                      &      2 048 &      CPU & 1 809.63 \\
  a0.20.256                      &      4 096 &      CPU & 2 790.53 \\
  a0.20.256                      &      8 192 &      CPU & 4 811.88 \\
  a0.20.256                      &     16 384 &      CPU & 6 639.80 \\
  a0.20.256                      &     32 768 &      CPU & 7 940.65 \\
  a0.20.256                      &     65 536 &      CPU & 9 395.22 \\
  mobile.conv.avg.bin.40.128.512 &         16 &      CPU &     5.72 \\
  mobile.conv.avg.bin.40.128.512 &         32 &      CPU &    11.40 \\
  mobile.conv.avg.bin.40.128.512 &         64 &      CPU &    23.16 \\
  mobile.conv.avg.bin.40.128.512 &        128 &      CPU &    45.58 \\
  mobile.conv.avg.bin.40.128.512 &        256 &      CPU &    88.34 \\
  mobile.conv.avg.bin.40.128.512 &        512 &      CPU &   193.15 \\
  mobile.conv.avg.bin.40.128.512 &      1 024 &      CPU &   377.37 \\
  mobile.conv.avg.bin.40.128.512 &      2 048 &      CPU &   643.34 \\
  mobile.conv.avg.bin.40.128.512 &      4 096 &      CPU & 1 344.05 \\
  mobile.conv.avg.bin.40.128.512 &      8 192 &      CPU & 2 108.71 \\
  mobile.conv.avg.bin.40.128.512 &     16 384 &      CPU & 3 919.71 \\
  mobile.conv.avg.bin.40.128.512 &     32 768 &      CPU & 5 754.23 \\
  mobile.conv.avg.bin.40.128.512 &     65 536 &      CPU & 6 679.22 \\
  \end{tabular}
\end{table}

\subsection{Making the networks play}

I made a round robin tournament between some of the networks in order to compare their level of play. The tournament gives each network one second per move using a RTX 2080 Ti. It accounts for around 300 evaluations per move for the big networks and around 500 evaluations per move for the small networks. The large residual networks result in more playouts per seconds than the relatively large Mobile networks. The MCTS algorithm used is PUCT. In order to have diversity in the games played by the same networks I randomized the choice of moves. Each move is ranked by the number of evaluations that are below it in the PUCT tree. If the second best move has more than half the number of evaluations of the best move, it becomes a candidate for the move to be played. The engine chooses the second best move with probability 0.5 when the second best move is candidate, otherwise it plays the best move.

The results of the tournament are given in table \ref{tableRoundRobin}. The networks that play are networks trained on the Leela dataset. The network that has the best level of play is the large Mobile network with a weight of 1 on the binary cross entropy value loss. It is followed by the large Mobile network with a weight of 4 on the value loss. Even if the MSE loss is smaller with a weight of 4 on the value it does not result in stronger play. The small Mobile network is slightly better that the 20 blocks residual network. The worst network is the small residual one since it could not learn the value and due to that only plays according to its policy.

\begin{table}
  \centering
  \caption{Round robin tournament between networks trained on the Leela dataset.}
  \label{tableRoundRobin}
  \begin{tabular}{lrrrrrrrrrr}
  
  Network                                    &  Games & Winrate & $\sigma$\\
                                             &        &         & \\
  mobile.conv.avg.bin.40.128.512             &    252 & 0.754 & 0.027 \\
  mobile.conv.avg.bin.val4.40.512.128        &    252 & 0.710 & 0.029 \\
  mobile.conv.avg.bin.33.200.64              &    252 & 0.671 & 0.030 \\
  a0.20.256                                  &    252 & 0.591 & 0.031 \\
  mobile.conv.avg.bin.val4.33.200.64         &    252 & 0.575 & 0.031 \\
  a0.conv.avg.bin.val4.13.64                 &    252 & 0.377 & 0.031 \\
  a0.conv.avg.bin.13.64                      &    252 & 0.313 & 0.028 \\
  a0.8.66                                    &    252 & 0.008 & 0.006 \\
  \end{tabular}
\end{table}






I also made the mobile.conv.avg.bin.40.128.512 network play on KGS. It plays instantly using the best move of the policy. It reached a stable 5 dan ranking. It is better than my previous residual policy network which reached a 4 dan ranking \cite{Cazenave2018residual}.

\section{Conclusion}

Residual networks were compared to Mobile networks with a fully convolutional policy head and a global average pooling value head. For the Leela dataset composed of games played at a superhuman level by a strong engine Mobile networks are better than residual networks both for small and for large networks. They have a better accuracy and value error. They are also better when compared according to the number of parameters of the networks. A tournament between the different networks using a fixed time per move confirmed that Mobile networks play better than residual networks that use many more parameters.

\section*{Acknowledgment}

The experiments of this research were made using the Jean Zay supercomputer of the French Ministry of Research hosted by the CNRS. Tristan Cazenave is supported by the PRAIRIE institute.

\bibliographystyle{IEEEtran}
\bibliography{main}

\begin{IEEEbiography}[{\includegraphics[width=1in,height=1.25in,clip,keepaspectratio]{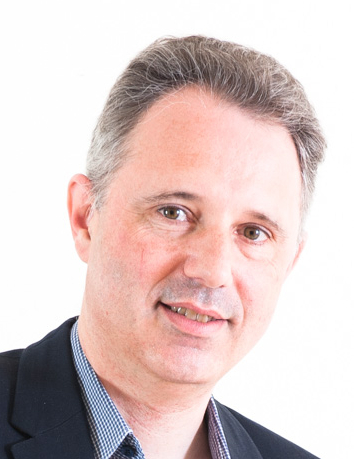}}]{Tristan Cazenave}
Professor of Artificial Intelligence at LAMSADE, University Paris-Dauphine, PSL Research University and CNRS. Author of more than a hundred scientific papers about Artificial Intelligence in games. He started publishing commercial video games at the age of 16 and defended a PhD thesis on machine learning for computer Go in 1996 at Sorbonne University.
\end{IEEEbiography}
\onecolumn 
\appendix[Source code] 

\begin{verbatim}
filters = 512
trunk = 128

def bottleneck_block(x, expand=filters, squeeze=trunk):
  m = layers.Conv2D(expand, (1,1), 
                    kernel_regularizer=regularizers.l2(0.0001), 
                    use_bias = False)(x)
  m = layers.BatchNormalization()(m)
  m = layers.Activation('relu')(m)
  m = layers.DepthwiseConv2D((3,3), padding='same', 
                             kernel_regularizer=regularizers.l2(0.0001), 
                             use_bias = False)(m)
  m = layers.BatchNormalization()(m)
  m = layers.Activation('relu')(m)
  m = layers.Conv2D(squeeze, (1,1), 
                    kernel_regularizer=regularizers.l2(0.0001), 
                    use_bias = False)(m)
  m = layers.BatchNormalization()(m)
  return layers.Add()([m, x])

def getModel ():
    input = keras.Input(shape=(19, 19, 21), name='board')
    x = layers.Conv2D(trunk, 1, padding='same', 
                      kernel_regularizer=regularizers.l2(0.0001))(input)
    x = layers.BatchNormalization()(x)
    x = layers.ReLU()(x)
    for i in range (blocks):
        x = bottleneck_block (x, filters, trunk)
    policy_head = layers.Conv2D(1, 1, activation='relu', padding='same', 
                                use_bias = False, 
                                kernel_regularizer=regularizers.l2(0.0001))(x)
    policy_head = layers.Flatten()(policy_head)
    policy_head = layers.Activation('softmax', name='policy')(policy_head)
    value_head = layers.GlobalAveragePooling2D()(x)
    value_head = layers.Dense(50, activation='relu', 
                              kernel_regularizer=regularizers.l2(0.0001))(value_head)
    value_head = layers.Dense(1, activation='sigmoid', name='value', 
                              kernel_regularizer=regularizers.l2(0.0001))(value_head)

    model = keras.Model(inputs=input, outputs=[policy_head, value_head])

    return model
\end{verbatim}

\end{document}